\documentclass{article}

\usepackage{microtype}
\usepackage{graphicx}
\usepackage{booktabs} 
\usepackage[sort&compress,numbers]{natbib}

\usepackage[accepted]{icml2024}
\newcommand{\murdochetal}{a prior work~\cite{murdoch2019Definitions}}
\newcommand{\lengerichetal}{The authors}
\usepackage{xcolor}
\definecolor{cblue}{RGB}{8, 85, 153}
\usepackage[citecolor=cblue, linkcolor=cblue,colorlinks=true, urlcolor=cblue, hyperfootnotes=false]{hyperref}
\renewcommand{\cite}[1]{\hspace{-0.40em}\textsuperscript{ \footnotesize \raisebox{-1.5pt}{\citenum{#1}}}}

\usepackage[hyperpageref]{backref}
\renewcommand*{\backref}[1]{\ifx#1\relax \else Page #1 \fi}
\renewcommand*{\backrefalt}[4]{%
    \ifcase #1 \scriptsize{(Not cited.)}%
    \or        \scriptsize{\textcolor{gray}{$\hookrightarrow$}#2.}%
    \else      \scriptsize{\textcolor{gray}{$\hookrightarrow$}#2}%
    \fi}

\makeatletter 
\renewcommand{\ICML@appearing}{%
}
\makeatother 

\usepackage{amsmath}
\usepackage{amssymb}
\usepackage{mathtools}
\usepackage{amsthm}

\usepackage[capitalize]{cleveref}
\usepackage[textsize=tiny]{todonotes}
\crefname{section}{Sec.}{Secs.}

\usepackage[labelfont=bf]{caption}
\usepackage{placeins}

\icmltitlerunning{Rethinking Interpretability in the Era of LLMs}

\begin{document}

\twocolumn[
\icmltitle{Rethinking Interpretability in the Era of Large Language Models}

\author{Chandan Singh, Jeevana Priya Inala, Michel Galley, Rich Caruana, Jianfeng Gao\\\\ \addr Microsoft Research}

\begin{icmlauthorlist}
\icmlauthor{Chandan Singh}{xxx}
\icmlauthor{Jeevana Priya Inala}{xxx}
\icmlauthor{Michel Galley}{xxx}
\icmlauthor{Rich Caruana}{xxx}
\icmlauthor{Jianfeng Gao}{xxx}
\end{icmlauthorlist}

\icmlaffiliation{xxx}{Microsoft Research}

\icmlcorrespondingauthor{Chandan Singh}{chansingh@microsoft.com}
\icmlkeywords{Interpretability, LLM, Explainability, XAI, Transparent ML}

\vskip 0.3in
]

\printAffiliationsAndNotice{}


\begin{abstract}
Interpretable machine learning has exploded as an area of interest over the last decade, sparked by the rise of increasingly large datasets and deep neural networks. Simultaneously, large language models (LLMs) have demonstrated remarkable capabilities across a wide array of tasks, offering a chance to rethink opportunities in interpretable machine learning. Notably, the capability to explain in natural language allows LLMs to expand the scale and complexity of patterns that can be given to a human. However, these new capabilities raise new challenges, such as hallucinated explanations and immense computational costs.

In this position paper, we start by reviewing existing methods to evaluate the emerging field of LLM interpretation (both interpreting LLMs and using LLMs for explanation). We contend that, despite their limitations, LLMs hold the opportunity to redefine interpretability with a more ambitious scope across many applications, including in auditing LLMs themselves. We highlight two emerging research priorities for LLM interpretation: using LLMs to directly analyze new datasets and to generate interactive explanations.
\end{abstract}

\section{Introduction}

Machine learning (ML) and natural language processing (NLP) have seen a rapid expansion in recent years, due to the availability of increasingly large datasets and powerful neural network models.
In response, the field of interpretable ML\footnote{We use the terms interpretable, explainable, and transparent interchangeably.} has grown to incorporate a diverse array of techniques and methods for understanding these models and datasets~\cite{doshi2017roadmap,murdoch2019Definitions,molnar2019interpretable}.
One part of this expansion has focused on the development and use of inherently interpretable models~\cite{rudin2021interpretable}, such as sparse linear models, generalized additive models, and decision trees.
Alongside these models, post-hoc interpretability techniques have become increasingly prominent, offering insights into predictions after a model has been trained. 
Notable examples include methods for assessing feature importance~\cite{ribeiro2016should,lundberg2017unified}, and broader post-hoc techniques, e.g., model visualizations~\cite{yosinski2015understanding,bau2018gan},
or interpretable distillation~\cite{tan2018distill,ha2021adaptive}. 

Meanwhile, pre-trained large language models (LLMs) have shown impressive proficiency in a range of complex NLP tasks, significantly advancing the field and opening new frontiers for applications~\cite{brown2020language,touvron2023llama2,openai2023gpt4}.
However, the inability to effectively interpret these models has debilitated their use in high-stakes applications such as medicine and raised issues related to
regulatory pressure, safety, and alignment~\cite{goodman2016european,amodei2016concrete,gabriel2020artificial}.
Moreover, this lack of interpretability has limited the use of LLMs (and other neural-network models) in fields such as science and data analysis~\cite{wang2023scientific,kasneci2023chatgpt,ziems2023can}.
In these settings, the end goal is often to elicit a trustworthy interpretation, rather than to deploy an LLM.

In this work, we contend that LLMs hold the opportunity to rethink interpretability with a more ambitious scope.
LLMs can elicit more elaborate explanations
than the previous generation of interpretable ML techniques.
While previous methods have often relied on restricted interfaces such as saliency maps, LLMs can communicate directly in expressive natural language.
This allows users to make targeted queries, such as \textit{Can you explain your logic?}, \textit{Why didn't you answer with (A)?}, or \textit{Explain this data to me.},
and get immediate, relevant responses.
We believe simple questions such as these,
coupled with techniques for grounding and processing data,
will allow LLMs to articulate previously incomprehensible model behaviors and data patterns directly to humans in understandable text.
However, unlocking these opportunities requires tackling new challenges,
including hallucinated (i.e. incorrect or baseless) explanations,
along with
the immense size, cost, and inherent opaqueness of modern LLMs.

\begin{figure*}[ht]
  \centering
  \hspace{150pt}\includegraphics[width=0.9\textwidth]{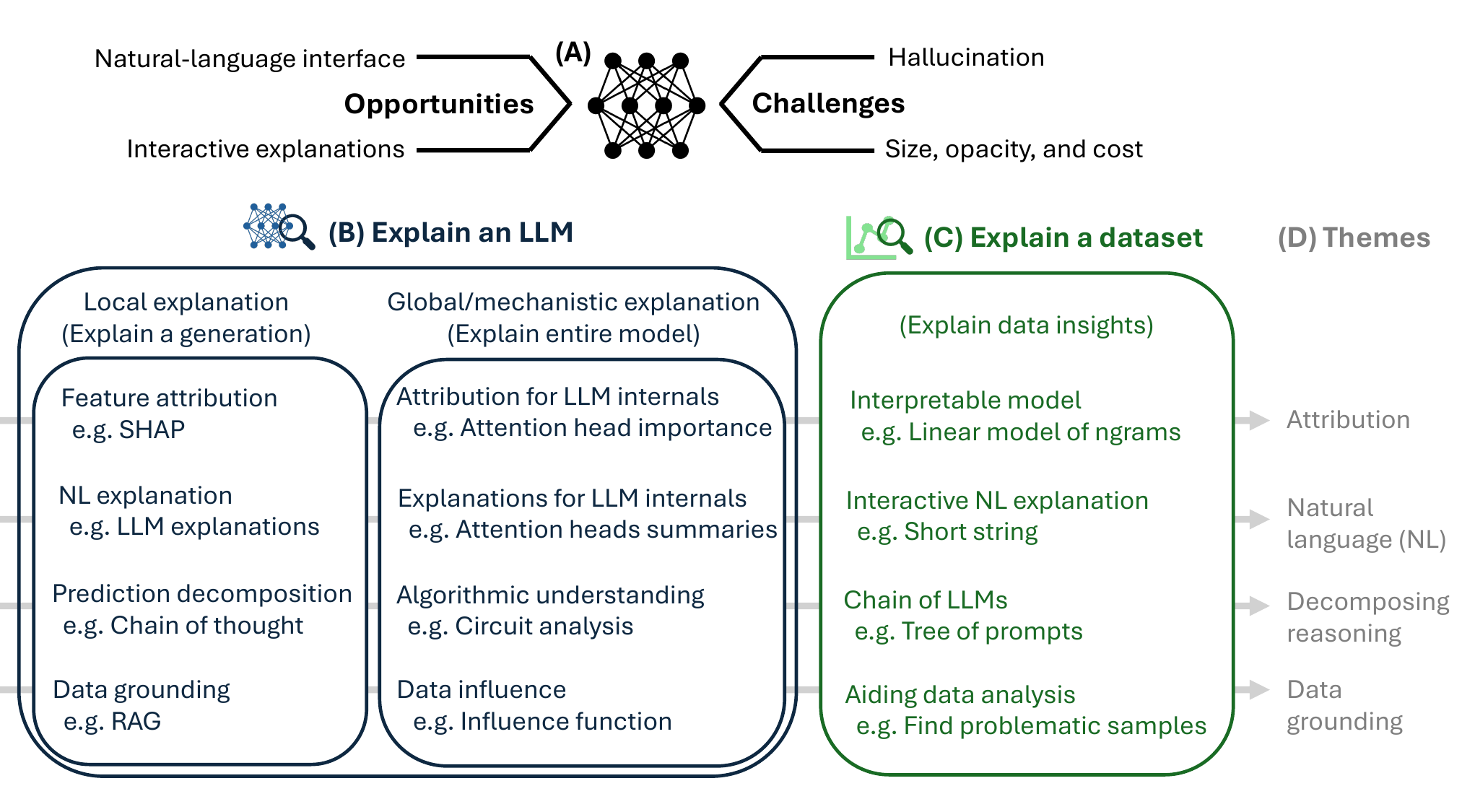}
  \caption{\textbf{Categorization of LLM interpretation research}.
\textbf{(A)} LLMs raise unique opportunities and challenges for interpretation (\cref{sec:unique_opportunities_challenges}).
\textbf{(B)} Explaining an LLM can be categorized into methods that seek to explain a single generation from an LLM (i.e. local explanation, \cref{subsec:local}) or the LLM in its entirety (i.e. global/mechanistic explanation, \cref{subsec:global}).
Local explanation methods build on many techniques that were originally developed for interpreting non-LLM models, such as feature attribution methods.
More recent local explanation techniques use LLMs themselves to yield interpretations,
e.g., through post-hoc natural language (NL) explanations, 
asking an LLM to build explanations into its generation process,
or through data grounding.
Similar techniques have been developed and applied to global explanation, although it also includes unique types of explanations, e.g., analyzing individual attention heads or circuits inside an LLM. 
\textbf{(C)} \cref{sec:data} analyzes the emerging area that uses an LLM to aid in directly explaining a \textit{dataset}.
In this setting, an LLM is given a new dataset (which can consist of either text or tabular features) and is used to help analyze it.
LLM-based techniques for dataset explanation are quite diverse, including helping to build interpretable models, generate NL explanations,
generate chains of NL explanations,
or construct data visualizations.
\textbf{(D)} Common themes emerge among methods for local explanation, global explanation, and dataset explanation.
}
  \label{fig:intro}
\end{figure*}

\paragraph{Contributions and overview}
We evaluate LLM interpretation and highlight emerging research priorities,
taking a broader scope than recent works, e.g., those focused on
explaining LLM predictions~\cite{zhao2023explainability}, mechanistic interpretability~\cite{rauker2023toward},
social science~\cite{ziems2023can},
or science more generally~\cite{wang2023scientific,birhane2023science,pion2021learning}.
Rather than providing an exhaustive overview of methods,
we highlight the aspects of interpretability that are unique to LLMs and showcase them with practically useful methods.

Specifically, we begin with a background and definitions~(\cref{sec:background}) before proceeding to analyze the unique opportunities and challenges that LLMs present for interpretation~(\cref{sec:unique_opportunities_challenges}).
We then ground these opportunities in two complementary categories for LLM-based interpretation (see \cref{fig:intro}).
The first is generating explanations for \textit{an existing LLM} (\cref{sec:model}), which is useful for auditing a model's performance, alignment, fairness, etc.
The second is explaining \textit{a dataset} (\cref{sec:data});
in this setting, an LLM is used to help analyze a new dataset (which can consist of either text or tabular features).

Throughout the paper, we highlight dataset explanation and interactive explanation as emerging research priorities.
Together, these two areas have great potential real-world significance in domains from science to statistics, where they can facilitate the process of scientific discovery, data analysis, and model building.
Throughout, we focus on pre-trained LLMs, mostly applied to text data, but also applied to tabular data.

\section{Background: definitions and evaluation}
\label{sec:background}
\paragraph{Definitions}

Without context, \textit{interpretability} is a poorly defined concept.
Precisely defining interpretability requires understanding the problem and audience an interpretation is intended to serve.
In light of this imprecision, interpretable ML has largely become associated with a narrow set of techniques, including feature attribution, saliency maps, and transparent models.
However, LLM interpretation is broader in scope and more expressive than these methods.
Here, we paraphrase the definition of interpretable ML from \murdochetal{} to define LLM interpretation as the \textit{extraction of relevant knowledge from an LLM concerning relationships either contained in data or learned by the model}.
We emphasize that this definition applies to both interpreting an LLM and to using an LLM to generate explanations.
Moreover, the definition relies on the extraction of \textit{relevant} knowledge, i.e., knowledge that is useful for a particular problem and audience.
For example, in a code generation context, a relevant interpretation may help a user quickly integrate an LLM-generated code snippet.
In contrast, a relevant interpretation in a medical diagnosis setting may inform a user whether or not a prediction is trustworthy.

The term \textit{large language model (LLM)} is often used imprecisely.
Here, we use it to refer to transformer-based neural language models that contain tens to hundreds of billions of parameters,
and which are pre-trained on massive text data, e.g., PaLM~\cite{chowdhery2023palm}, LLaMA~\cite{touvron2023llama2}, and GPT-4~\cite{openai2023gpt4}.
Compared to early pre-trained language models, such as BERT, LLMs are not only much larger, but also exhibit stronger language understanding, generation abilities, and explanation capabilities.
After an initial computationally intensive pre-training stage, LLMs often undergo instruction finetuning and further alignment with human preferences to improve instruction following~\cite{ouyang2022training} or to improve interactive chat capabilities, e.g., the LLaMA-2 chat model~\cite{touvron2023llama2}.
They are sometimes also further adapted via supervised finetuning to improve performance in a specific domain, such as medicine~\cite{singhal2023towards}.

After undergoing these steps, LLMs are often used with \textit{prompting}, the most common interface for applying LLMs (and our main focus in this paper).
In prompting, a text prompt is directly fed to an LLM and used to generate subsequent output text.
\textit{Few-shot prompting} is a type of prompting that involves providing an LLM with a small number of examples to allow it to better understand the task it is being asked to perform.

\paragraph{Evaluating LLM interpretations}

Since different interpretations are relevant to different contexts, the ideal way to evaluate an interpretation is by studying whether its usage in a real-world setting with humans improves a desired outcome~\cite{kim2017interpretability}.
In contrast, simply measuring human judgment of explanations is not particularly useful, as it may not translate into improvements in practice~\cite{adebayo2018sanity}.
A recent meta-analysis finds that introducing NLP explanations into settings with humans yields widely varying utilities, ranging from completely unhelpful to very useful~\cite{chaleshtori2023evaluating}.
An important piece of this evaluation is the notion of complementarity~\cite{bansal2021does}, i.e., that explanations should help LLMs complement human performance in a team setting, rather than improve their performance in isolation.

While human studies provide the most realistic evaluation, automated metrics (that can be computed without involving humans) are desirable to ease and scale evaluation, especially in mechanistic interpretability.
An increasingly popular approach is to use LLMs themselves in evaluation, although great care must be taken to avoid introducing biases, e.g., an LLM systematically scoring its own outputs too positively~\cite{zheng2023judging}.
One way to reduce bias is to use LLMs as part of a structured evaluation process tailored to a particular problem, rather than directly querying LLMs for evaluation scores.
For example, one common setting is evaluating a natural-language interpretation of a given function (which may be any component of a pre-trained LLM).
In this setting, one can evaluate an explanation's ability to simulate the function's behavior~\cite{bills2023language}, the function's output on LLM-generated synthetic data~\cite{singh2023explainingmodules},
or its ability to recover a groundtruth function~\cite{schwettmann2023find,zhong2023goaldd}.
In a question-answering setting, many automated metrics have been proposed for measuring the faithfulness of a natural-language explanation for an individual answer to a question~\cite{atanasova2023faithfulness,parcalabescu2023measuring,chen2022rev}.

A final avenue for evaluating interpretations is through their ability to alter/improve model performance in useful ways.
This approach provides strong evidence for the utility of an explanation, although it does not encompass all critical use cases of interpretability (particularly those directly involving human interaction).
Model improvements can take various forms, the simplest of which is simply improving accuracy at downstream tasks.
For example,
few-shot accuracy was seen to improve when
aligning an LLM's rationales with explanations generated using post-hoc explanation methods~\cite{krishna2023post} or explanations distilled from large models~\cite{mukherjee2023orca}.
Moreover, employing few-shot explanations during inference (not training) can significantly improve few-shot LLM accuracy, especially when these explanations are further optimized~\cite{lampinen2022can,ye2023explanation}.
Beyond general performance, explanations can be used to overcome specific shortcomings of a model.
For example, one line of work identifies and addresses shortcuts/spurious correlations learned by an LLM~\cite{du2023shortcut,kang2023impact,bastings2021will}.
Model editing, a related line of work, enables precise modifications to certain model behaviors, enhancing overall performance~\cite{meng2022locating,mitchell2022fast,hernandez2023inspecting}.

\section{Unique opportunities and challenges of LLM interpretation}
\label{sec:unique_opportunities_challenges}

\paragraph{Unique opportunities of LLM interpretation}

First among LLM interpretation opportunities is the ability to provide \textit{a natural-language interface} to explain complex patterns.
This interface is very familiar to humans, potentially ameliorating the difficulties that practitioners often face when using explainability techniques~\cite{kaur2020interpreting,weld2019challenge}.
Additionally, natural language can be used to build a bridge between humans and a range of other modalities, e.g., DNA, chemical compounds, or images~\cite{taylor2022galactica,Liu2022MultimodalMS,radford2021learning},
that may be difficult for humans to interpret on their own.
In these cases, natural language allows for expressing complex concepts through explanations at different levels of granularity, potentially grounded in evidence or discussions of counterfactuals.
 
A second major opportunity is the ability for LLMs to generate \textit{interactive explanations}.
Interactivity allows users to tailor explanations to their unique needs, e.g., by asking follow-up questions and performing analysis on related examples.
Interviews with decision-makers, including physicians and policymakers, indicate that they strongly prefer interactive explanations, particularly in the form of natural-language dialogues~\cite{lakkaraju2022rethinking}.
Interactivity further allows LLM explanations to be decomposed into many different LLM calls, each of which can be audited independently.
This can be enabled in different ways, e.g., having a user repeatedly chat with an LLM using prompting,
or providing a user a sequence of LLM calls and evidence to analyze.

\paragraph{Unique challenges of LLM interpretation}
These opportunities bring new challenges.
First and foremost is the issue of \textit{hallucination}, i.e. incorrect or baseless explanations.
Flexible explanations provided in natural language can quickly become less grounded in evidence, whether the evidence is present in a given input or presumed to be present in the knowledge an LLM has learned from its training data.
Hallucinated explanations are unhelpful or even misleading, and thus techniques for identifying and combating hallucination are critical to the success of LLM interpretation.

A second challenge is the \textit{immensity and opaqueness} of LLMs.
Models have grown to contain tens or hundreds of billions of parameters~\cite{brown2020language,touvron2023llama2}, and continue to grow in size.
This makes it infeasible for a human to inspect or even comprehend the units of an LLM.
Moreover, it necessitates efficient algorithms for interpretation, as even generating a single token from an LLM often incurs a non-trivial computational cost.
In fact, LLMs are often too large to be run locally or can be accessed only through a proprietary text API, necessitating the need for interpretation algorithms that do not have full access to the model (e.g., no access to the model weights or the model gradients).

\section{Explaining an LLM}
\label{sec:model}

In this section, we study techniques for explaining an LLM, including explaining a single generation from an LLM (\cref{subsec:local}) or an LLM in its entirety (\cref{subsec:global}).
We evaluate both traditional interpretable ML techniques and LLM-based techniques for explaining an LLM.

\subsection{Local explanation}
\label{subsec:local}
Local explanation, i.e., explaining a single generation from an LLM, has been a major focus in the recent interpretability literature.
It allows for understanding and using LLMs in high-stakes scenarios, e.g., healthcare.

The simplest approach for providing local explanations in LLMs provides feature attributions for input tokens.
These feature attributions assign a relevance score to each input feature, reflecting its impact on the model's generated output.
Various attribution methods have been developed, including perturbation-based methods~\cite{lundberg2017unified}, gradient-based methods~\cite{sundararajan2016gradients,montavon2017explaining}, and linear approximations~\cite{ribeiro2016should}.
Recently, these methods have been specifically adapted for transformer models, addressing unique challenges such as discrete token embeddings~\cite{sikdar2021integrated,enguehard2023sequential} and computational costs~\cite{chen2023algorithms}.
Moreover, the conditional distribution learned by an LLM can be used to enhance existing attribution methods, e.g., by performing input marginalization~\cite{kim2020interpretation}.
Besides feature attributions, attention mechanisms within an LLM offer another avenue for visualizing token contributions to an LLM generation~\cite{wiegreffe2019attention}, though their faithfulness/effectiveness remains unclear~\cite{jain2019attention}.
Interestingly, recent work suggests that LLMs themselves can generate post-hoc attributions of important features through prompting~\cite{kroeger2023large}.
This approach could be extended to enable eliciting different feature attributions that are relevant in different contexts.

Beyond token-level attributions, LLMs can also generate local explanations directly in natural language.
While the generation of natural-language explanations predates the current era of LLMs (e.g., in text classification~\cite{camburu2018snli,rajani2019explain} or image classification~\cite{hendricks2016generating}),
the advent of more powerful models has significantly enhanced their effectiveness.
Natural-language explanations generated by LLMs have shown the ability to elucidate model predictions, even simulating counterfactual scenarios~\cite{bhattacharjee2023llms}, and expressing nuances like uncertainty~\cite{xiong2023can,tanneru2023quantifying,zhou2024relying}.
Despite their potential benefits, natural language explanations remain extremely susceptible to hallucination or inaccuracies, especially when generated post-hoc~\cite{chen2023models,ye2022unreliability}.

One starting point for combating these hallucinations is integrating an explanation within the answer-generation process itself.
Chain-of-thought prompting exemplifies this approach~\cite{wei2022chain}, where an LLM is prompted to articulate its reasoning step-by-step before arriving at an answer.
This reasoning chain generally results in more accurate and faithful outcomes, as the final answer is more aligned with the preceding logical steps. The robustness of this method can be tested by introducing perturbations in the reasoning process and observing the effects on the final output~\cite{madaan2022text,wang2022towards,Lanham2023MeasuringFI}.
Alternative methods for generating this reasoning chain exist, such as tree-of-thoughts~\cite{yao2023tree},
which extends chain-of-thought to instead generate a tree of thoughts used in conjunction with backtracking,
graph-of-thoughts~\cite{besta2023graph},
and others~\cite{Nye2021ShowYW,press2022measuring,zhou2022least}.
All of these methods not only help convey an LLM's intermediate reasoning to a user,
but also help the LLM to follow the reasoning through prompting, often enhancing the reliability of the output.
However, like all LLM-based generations, the fidelity of these explanations can vary~\cite{Lanham2023MeasuringFI,wei2023larger}. 

An alternative path to reducing hallucinations during generation is to employ retrieval-augmented generation (RAG).
In RAG, an LLM incorporates a retrieval step in its decision-making process,
usually by searching a reference corpus or knowledge base using text embeddings~\cite{Guu2020REALMRL,Peng2023CheckYF} (see review~\cite{worledge2023unifying}).
This allows the information that is used to generate an output to be specified and examined explicitly, making it easier to explain the evidence an LLM uses during decision-making.

\subsection{Global and mechanistic explanation}
\label{subsec:global}

Rather than studying individual generations,
global / mechanistic explanations aim to understand an LLM as a whole.
These explanations can help to audit a model for concerns beyond generalization, e.g., bias, privacy, and safety, helping to build LLMs that are more efficient /
trustworthy,
They can also yield mechanistic understanding about how LLMs function.
To do so, researchers have focused on summarizing the behaviors and mechanisms of LLMs through various lenses.
Generally, these works require access to model weights and do not work for explaining models that are only accessible through a text API, e.g., GPT-4~\cite{openai2023gpt4}.

One popular method for understanding neural-network representations is probing.
Probing techniques analyze a model's representation either by decoding embedded information, e.g., syntax~\cite{conneau2018you}, or by testing the model's capabilities through precisely designed tasks, e.g., subject-verb agreement~\cite{liu2019incorporating,marvin2018targeted}. 
In the context of LLMs, probing has evolved to include the analysis of attention heads~\cite{clark2019does}, embeddings~\cite{morris2023text}, and different controllable aspects of representations~\cite{zou2023RepresentationEA}.
It also includes methods that directly decode an output token to understand what is represented at different positions and layers~\cite{belrose2023eliciting,ghandeharioun2024patchscope}.
These methods can provide a deeper understanding of the nuanced ways in which LLMs process and represent information.

In addition to probing, many works study LLM representations at a more granular level.
This includes categorizing or decoding concepts from individual neurons~\cite{mu2020compositional,gurnee2023finding}
or directly explaining the function of attention heads in natural language~\cite{bills2023language,singh2023explainingmodules,hernandez2022natural}.
Beyond individual neurons,
there is growing interest in understanding how groups of neurons combine to perform specific tasks,
e.g., finding a circuit for indirect object identification~\cite{wang2022interpretability},
for entity binding~\cite{feng2023language},
or for multiple shared purposes~\cite{merullo2023circuit}.
More broadly, this type of analysis can be applied to localize functionalities rather than fully explain a circuit, e.g., localizing factual knowledge within an LLM~\cite{meng2022locating,dai2021knowledge}.
A persistent problem with these methods is that they are difficult to scale to immense LLMs, leading to research in (semi)-automated methods that can scale to today's largest LLMs~\cite{lieberum2023does,Wu2023InterpretabilityAS}.

A complementary approach to mechanistic understanding uses miniature LLMs as a test bed for investigating complex phenomena.
For example, examining a 2-layer transformer model reveals information about what patterns are learned by attention heads as a function of input statistics~\cite{elhage2021mathematical} or helps identify key components, such as induction heads or ngram heads that copy and utilize relevant tokens~\cite{olsson2022context,akyürek2024incontext}.
This line of mechanistic understanding places a particular focus on studying the important capability of in-context learning, i.e., given a few input-output examples in a prompt, an LLM can learn to correctly generate an output for a new input~\cite{garg2022can,zhou2023algorithms}.

A related area of research seeks to interpret an LLM by understanding the influence of its training data distribution.
Unlike other methods we have discussed, this requires access to an LLM's training dataset, which is often unknown or inaccessible.
In the case that the data is known,
researchers can employ techniques such as influence functions to identify important elements in the training data~\cite{grosse2023studying}.
They can also study how model behaviors arise from patterns in training data,
such as hallucination in the presence of long-tail data~\cite{kandpal2023large},
in the presence of repeated training data~\cite{hernandez2022scaling},
or statistical patterns that contradict proper reasoning~\cite{mckenna2023sources}.

All these interpretation techniques can be improved via LLM-based interactivity, allowing a user to investigate different model components via follow-up queries and altered prompts from a user.
For example, one recent work introduces an end-to-end framework for explanation-based debugging and improvement of text models, showing that it can quickly yield improvements in text-classification performance~\cite{lee2022xmd}.
Another work, Talk2Model, introduces a natural-language interface that allows users to interrogate a tabular prediction model through a dialog, implicitly calling many different model explainability tools, such as calculating feature importance~\cite{slack2022talktomodel}.\footnote{Note that Talk2Model focuses on interpreting prediction models rather than LLMs.}
More recent work extends Talk2Model to a setting interrogating an LLM about its behavior~\cite{wang2024llmcheckup}.

Finally, the insights gained from mechanistic understanding are beginning to inform practical applications, with current areas of focus including model editing~\cite{meng2022locating}, improving instruction following~\cite{zhang2023tell}, and model compression~\cite{sharma2023truth}.
These areas simultaneously serve as a sanity check on many mechanistic interpretations and as a useful path to enhancing the reliability of LLMs.

\section{Explaining a dataset}
\label{sec:data}

\begin{figure*}
    \centering
    \includegraphics[width=1.0\textwidth]{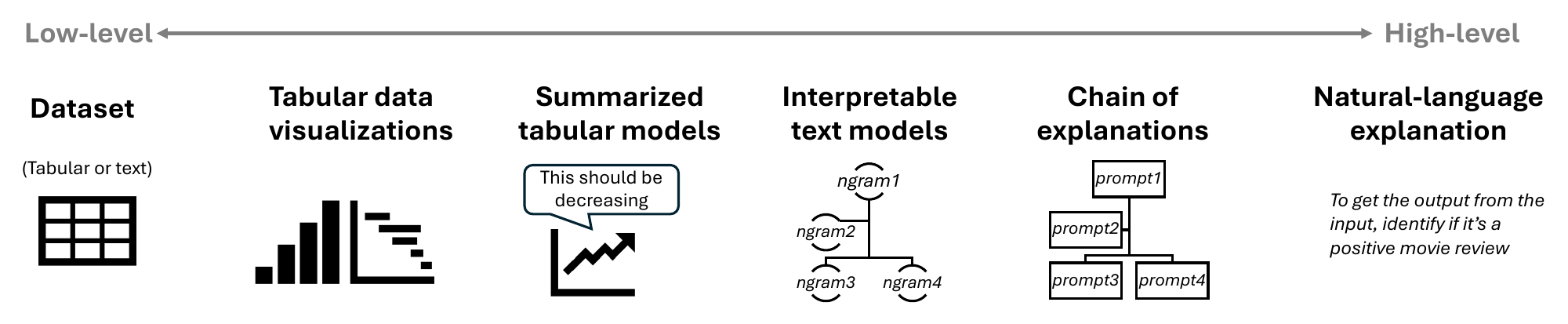}
    \vspace{-10pt}
    \caption{\textbf{Dataset explanations at different levels of granularity.}
    Dataset explanation involves understanding a new dataset (consisting of either text or tabular features) using a pre-trained LLM.
    Low-level explanations are more faithful to the dataset but involve more human effort to extract meaningful insights.
    Many dataset interpretations use prediction models (classification or regression) as a means to identify and explain patterns between features.
    }
    \label{fig:dataset_explanation}
\end{figure*}

As LLMs improve their context length and capabilities,
they can be leveraged to explain an entire dataset, rather than explaining an LLM or its generations.
This can aid with data analysis, knowledge discovery, and scientific applications.
\cref{fig:dataset_explanation} shows an overview of dataset explanations at different levels of granularity, which we cover in detail below.
We distinguish between tabular and text data, but note that most methods can be successfully applied to either, or both simultaneously in a multimodal setting.

\paragraph{Tabular data}
One way LLMs can aid in dataset explanation is by making it easier to interactively visualize and analyze tabular data.
This is made possible by the fact that LLMs can simultaneously understand code, text, and numbers by treating them all as input tokens.
Perhaps the most popular method in this category is ChatGPT Code Interpreter\footnote{\url{https://openai.com/blog/chatgpt-plugins\#code-interpreter}}, which enables uploading datasets and building visualizations on top of them through an interactive text interface.
This capability is part of a broader trend of LLM-aided visualization, e.g., suggesting automatic visualizations for dataframes~\cite{dibia2023lida}, helping to automate data wrangling~\cite{narayan2022can}, or even conducting full-fledged data analysis~\cite{huang2023benchmarking}.
These capabilities benefit from a growing line of work that analyzes how to effectively represent and process tabular data with LLMs~\cite{li2023table,zhang2023towards,zhang2023generative}.

LLMs can also help explaining datasets by directly analyzing models that have been fit to tabular data
Unlike mechanistic interpretability,
where the goal is to understand the model,
in dataset explanation, the goal is to understand patterns in the data through the model (although similar techniques can be used for both problems).
For example, one recent work uses LLMs to analyze generalized additive models (GAMs) that are fit to tabular data~\cite{lengerich2023llms}.
GAMs are interpretable models that can be represented as a set of curves,
each representing the contribution of a feature to the output prediction as a function of the feature's value.
An LLM can analyze the fitted model (and thereby the underlying dataset) by processing each curve as a set of numerical tokens and then detecting and describing patterns in each curve.
\lengerichetal{} find that LLMs can identify surprising characteristics in the curves and the underlying data, largely based on their prior knowledge of a domain.
Rather than using an interpretable GAM model,
another approach is to distill dataset insights by analyzing classifier predictions.
For example, MaNtLE generates natural-language descriptions of a classifier's rationale based on the classifier's predictions,
and these explanations are found to identify explainable subgroups that contain similar feature patterns~\cite{menon2023mantle}.

\paragraph{Text data}

Text data poses different challenges for dataset explanation than tabular data because it is sparse, high-dimensional, and modeling it requires many high-order interactions.
As a result,
interpretable models that have been successful in the tabular domain (e.g., sparse linear models~\cite{tibshirani1996regression,ustun2016supersparse}, GAMs~\cite{hastie1986generalized,lou2013accurate,caruana2015intelligible}, decision trees~\cite{breiman1984classification,quinlan1986induction,agarwal2022Hierarchical}, and others~\cite{singh2021imodels,tan2022Fast}),
have struggled to accurately model text.
One recent line of work addresses this issue by using LLMs to help build fully interpretable text models, such as linear models or decision trees~\cite{singh2023augmenting};
the resulting models are surprisingly accurate, often outperforming even much larger LLM models.
These interpretable models can help explain a dataset by showing which features (i.e. words or ngrams) are important for predicting different outcomes.
Similar methods, e.g., CHiLL~\cite{mcinerney2023chill} use LLMs to build interpretable representations for text classification tasks.

Going beyond fully interpretable models, LLMs also help in building partially interpretable text models.
Partially interpretable text models often employ chains of prompts;
these chains allow for decomposing an LLM's decision-making process to analyze which dataset patterns a model learns.
Prompt chains are usually constructed by humans or by querying a model to generate a chain of calls on-the-fly~\cite{grunde2023designing}.
For dataset explanation, the most relevant chains are sequences of explanations that are generated by an LLM.
For example, a model can generate a single tree of explanations that is shared across all examples in a dataset, a process that enables understanding hierarchical structures stored within a dataset~\cite{morris2023tree}.
Rather than a tree, a single chain of prompts can often help an LLM employ self-verification, i.e. the model itself checks its previous generations using a chain of prompts, a popular technique that often improves reliability~\cite{pan2023automatically,madaan2023selfrefine,gero2023self}.
As in local explanation, an LLM can incorporate a retrieval step in its decision-making process~\cite{worledge2023unifying}, and access to different tools can help make different steps (e.g., arithmetic) more reliable and transparent~\cite{mialon2023augmented}.

Natural-language explanations hold the potential to produce rich, concise descriptions of patterns present in a dataset, but are prone to hallucination.
One method, iPrompt~\cite{singh2023explaining}, aims to avoid hallucination by searching for a dataset explanation in the form of a single prompt,
and verifying that the prompt induces an LLM to accurately predict a pattern in the underlying dataset.
Related methods use LLMs to provide descriptions that differentiate between groups in a dataset, followed by an LLM that verifies the credibility of the description~\cite{pmlr-v162-zhong22a,zhong2023goaldd,zhu2022gsclip}.
In addition to a raw natural-language explanation, LLMs can aid in summarizing textual information, e.g., through explainable clustering of a text dataset~\cite{wang2023goal} or creating prompt-based topic models~\cite{pham2023topicgpt}.

\section{Future research priorities}
\label{sec:discussion}


We now highlight research priorities surrounding LLM interpretation in three areas: explanation reliability, dataset explanation, and interactive explanations.

\paragraph{Explanation reliability}
All LLM explanations are bottlenecked by reliability issues.
This includes hallucinations~\cite{tonmoy2024comprehensive}, but encompasses a broader set of issues.
For example, LLMs continue to be very sensitive to the nuances of prompt phrasing; minor variations in prompts can completely change the substance of an LLM output~\cite{sclar2023quantifying,turpin2023language}.
Additionally, LLMs may ignore parts of their context, e.g., the middle of long contexts~\cite{Liu2023LostIT} or instructions that are difficult to parse~\cite{zhang2023tell}.

These reliability issues are particularly critical in interpretation, which often uses explanations to mitigate risk in high-stakes settings.
One work analyzing explanation reliably finds that LLMs often generate seemingly correct explanations that are actually inconsistent with their own outputs on related questions~\cite{chen2023models}, preventing a human practitioner from trusting an LLM or understanding how its explanations apply to new scenarios.
Another study finds that explanations generated by an LLM may not entail the model's predictions or be factually grounded in the input, even on simple tasks with extractive explanations~\cite{ye2022unreliability}.
Future work will be required to improve the grounding of explanations and develop stronger methods to test their reliability, perhaps through methods such as self-verification~\cite{pan2023automatically},
iterative prompting~\cite{singh2023explaining},
or automatically improving model self-consistency~\cite{chen2024consistent,li2023benchmarking,akyurek2024deductive}.

\paragraph{Dataset explanation for knowledge discovery}
Dataset explanation using LLMs (\cref{sec:data}) holds the potential 
to help with the generation and discovery of new knowledge from data~\cite{wang2023scientific,birhane2023science,pion2021learning}, rather than simply helping to speed up data analysis or visualization.
Dataset explanation could initially help at the level of brainstorming scientific hypotheses that can then be screened or tested by human researchers~\cite{yang2023large}.
During and after this process,
LLM explanations can help with
using natural language to understand data from otherwise opaque domains, such as chemical compounds~\cite{liu2023multi} or DNA sequences~\cite{taylor2022galactica}.
In the algorithms domain, LLMs have been used to uncover new algorithms, translating them to humans as readable computer programs~\cite{FunSearch2023}.
These approaches could be combined with data from experiments to help yield new data-driven insights.

LLM explanations can also be used to help humans better perform a task.
Explanations from transformers have already begun to be applied to domains such as Chess, where their explanations can help improve even expert players~\cite{schut2023bridging}.
Additionally, LLMs can provide explanations of expert human behavior, e.g. ``Why did the doctor prescribe this medication given this information about the patient?'', that are helpful in understanding, auditing, and improving human behavior~\cite{tu2024towards}.

\paragraph{Interactive explanations}
Finally, advancements in LLMs are poised to allow for the development of more user-centric, interactive explanations.
LLM explanations and follow-up questions are already being integrated into a variety of LLM applications, such as interactive task specification~\cite{li2023eliciting}, recommendation~\cite{huang2023recommender},
and a wide set of tasks involving dialog.
Furthermore, works like Talk2Model~\cite{slack2022talktomodel} enable users to interactively audit models in a conversational manner.
This dialog interface could be used in conjunction with many of the methods covered in this work to help with new applications, e.g., interactive dataset explanation.

\section{Conclusions}

In this paper, we have explored the vast and dynamic landscape of interpretable ML, particularly focusing on the unique opportunities and challenges presented by LLMs. LLMs' advanced natural language generation capabilities have opened new avenues for generating more elaborate and nuanced explanations, allowing for a deeper and more accessible understanding of complex patterns in data and model behaviors. As we navigate this terrain, we assert that the integration of LLMs into interpretative processes is not merely an enhancement of existing methodologies but a transformative shift that promises to redefine the boundaries of machine learning interpretability.


Our position is anchored in the belief that the future of interpretable ML hinges on our ability to harness the full potential of LLMs. To this end, we outlined several key stances and directions for future research, such as enhancing explanation reliability and advancing dataset interpretation for knowledge discovery.
As LLMs continue to improve rapidly, these explanations (and all the methods discussed in this work) will advance correspondingly to enable new applications and insights.
In the near future, LLMs may be able to offer the holy grail of interpretability:
explanations that can reliably aggregate and convey extremely complex information to us all.
\FloatBarrier
{
    \scriptsize
    \bibliographystyle{unsrt}

}

\end{document}